# GPT-4o Lacks Core Features of Theory of Mind

**John Muchovej (john.muchovej@yale.edu)[1], Amanda Royka[1], Shane Lee[2] & Julian Jara-Ettinger[1,2]**
[1] Department of Psychology, Yale University, New Haven, CT 06511
[2] Department of Computer Science, Yale University, New Haven, CT 06511

## Abstract

Do Large Language Models (LLMs) possess a Theory of Mind (ToM)? Research into this question has focused on evaluating LLMs against benchmarks and found success across a range of social tasks. However, these evaluations do not test for the actual representations posited by ToM: namely, a causal model of mental states and behavior. Here, we use a cognitively-grounded definition of ToM to develop and test a new evaluation framework. Specifically, our approach probes whether LLMs have a coherent, abstract, and consistent model of how mental states cause behavior – regardless of whether that model matches a human-like ToM. We find that even though LLMs succeed in approximating human judgments in a simple ToM paradigm, they fail at a logically equivalent task and exhibit low consistency between their action predictions and corresponding mental state inferences. As such, these findings suggest that the social proficiency exhibited by LLMs is not the result of an abstract or consistent ToM.

## Introduction

Do LLMs have a ToM? This question is not an idle philosophical inquiry. Instead, it is a crucial question for both cognitive science and artificial intelligence. For cognitive scientists, the possibility of LLM ToM represents an intriguing alien-like intelligence useful for testing theories ranging from the learnability of ToM to the relationship between ToM and capacities such as pragmatic reasoning. On the other hand, for artificial intelligence, LLM ToM would constitute a quantum leap forward for human-machine interaction. If LLMs had a ToM, we could be more confident that they will make reasonable, internally-consistent inferences about the social world.

However, research into LLM ToM has been inconclusive. Some work demonstrates remarkable successes (Gandhi et al., 2023; Kosinski, 2024; Moghaddam & Honey, 2023) while other investigations reveal striking fragility (Kim et al., 2023; Ma et al., 2023; Sap et al., 2022; Shapira et al., 2024; Trott et al., 2023; Ullman, 2023; Zhou et al., 2023). Yet, much of this work has focused on developmentally-inspired paradigms (Bubeck et al., 2023; Duijn et al., 2023; Strachan et al., 2024) raising an important question regarding construct validity: are these tests actually useful for probing ToM in LLMs, specifically? In the case of LLMs and their potential for ToM, we want to test whether they exhibit an emergent causal model that allows them to generalize outside of their training data. Thus, developmental studies, which are designed to control for low-level features and observable correlates of mental states, are not designed with LLM-related concerns in mind (e.g., ensuring that evaluations are sufficiently out-of-distribution to illustrate model generality). Thus, it is possible that LLMs exhibited high levels of social proficiency on past ToM tasks wholly in the absence of a ToM.

Given the slate of contradictory evidence, and the dangers of conflating social proficiency with genuine ToM, we offer a new proposal for testing LLM ToM. Inspired by developmental and computational cognition, our proposal moves away from traditional approaches that benchmark LLMs against human performance. Instead, we focus on the defining feature of ToM: namely, that it is a theory (Gopnik & Meltzoff, 1997, formally defined on p. 34-41).

ToM is considered a theory because it is a unified set of principles that can be used to both predict and explain a given phenomenon. Specifically, ToM explains the relationship between mental states and behavior through unified principles that take the form of a causal model of how mental-states generate behavior (Bora et al., 2009; Gopnik & Wellman, 1992; Heyes, 1998; Premack & Woodruff, 1978). Critically, since ToM is a theory, it possesses three defining characteristics: coherence, abstractness, and consistency. Here, we propose that these three characteristics can serve as desiderata when evaluating whether or not LLMs have a ToM. In the three studies below, we introduce evaluation methods for assessing each of these characteristics to arrive a more cognitively-grounded answer to the question, "Do LLMs have a ToM?".

## Study 1: Is LLM ToM coherent?

To begin, we tested whether LLMs have a coherent ToM. A coherent ToM would generate action-predictions by applying a set of core representations in a systematic way. For instance, the coherence of human ToM is derived from the principles of rational planning (Gergely & Csibra, 2003; Jara-Ettinger et al., 2016): representations of an action's costs and rewards are systematically combined with mental-state representations to generate predictions about an agent's likely behavior. Thus, as a first test of the theory-like nature of LLM "ToM," we begin by contrasting LLM action prediction across parametrically varying costs, desires, and beliefs, and compare those outputs to a model of human ToM (`HumanToM`) and more proto-human-like ToMs by lesioning `HumanToM` (Table 1B).

### Stimuli

We wanted a paradigm that could be fully enumerated, supporting a comprehensive evaluation of an LLM's forward model $\mathcal{F}$ of how representations of costs, desires, and beliefs combine to generate actions. Thus, we drew inspiration from the methods typically used to test ToM in computational models to design our *ContainerWorld* paradigm.

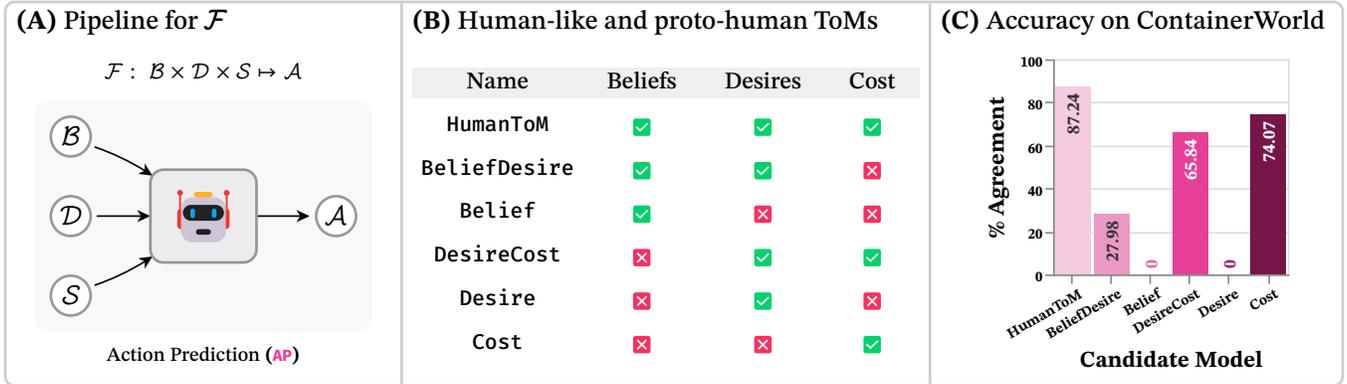

Figure 1: (A) For each set of beliefs $\mathcal{B}$, desires $\mathcal{D}$, and states $\mathcal{S}$ we queried an LLM for its action predictions $\mathcal{A}$. (B) The LLM's responses were compared against a set of candidate comparison models. Each candidate ToM model ablated different aspects of the full model to test whether LLMs systematically lack a particular aspect of ToM relative to humans' forward model. (C) GPT-4o's agreement (y-axis) compared with the candidate models (x-axis) listed in Table 1B.

**Paradigm: *ContainerWorld*** In this paradigm (Figure 2A), there is a character that spawns in the north-west corner of a room. This room has a closed box next to the character and a covered basket in the opposite corner "about 50 steps away". Each container holds fruit, which can be described as their state $\mathcal{S} \in$ {apples, oranges, apples and oranges}. While the contents of the containers are not directly observable to the character from their starting point, the character begins each round with beliefs about the contents of each container $\mathcal{B} \in$ {apples, oranges, apples and oranges}. The character's desires towards each fruit may be either $\mathcal{D} \in$ {like, dislike}, but they may not "dislike" both. Lastly, the character may take an action $\mathcal{A} \in$ {box, basket} – this entails moving to one container and taking one of the contents within.

**Procedure**

**Extracting the forward model from an LLM** To evaluate an LLM, we queried its $\mathcal{F}$ (Figure 1A) for action pre-dictions based on different *ContainerWorld* configurations. Specifically, we generated all possible tuples of beliefs $\mathcal{B}$, desires $\mathcal{D}$, and states $\mathcal{S}$ (there are 9 × 3 × 9 such tuples) to synthesize a parametrically-varying user prompt containing the transcription of $\langle \mathcal{B}, \mathcal{D}, \mathcal{S} \rangle$. The user prompt always concluded with a question prompting the LLM to generate its response ("Which container would Jason open?"). Additionally, all queries included the same system prompt informing the LLM of its task and providing a JSON schema informing the LLM how to structure its response (all system and user prompts are available on OSF). We then extracted the LLM's probability distribution over next tokens (e.g., "box" or "basket") and used this as the LLM's distribution over predicted next actions $\mathcal{A}$. The probability distribution here can be thought of as the LLM's confidence that the next token (word) should be "box" or "basket".

**Evaluation**

To evaluate the kind of ToM that an LLM may possess, we developed a series of candidate comparison models. These models capture human judgments by integrating beliefs, desires, and costs to compute the utility of different possible actions and generate a prediction (Jara-Ettinger et al., 2016). To this end, we can generate human-like, and proto-human-like, predictions within our *ContainerWorld* paradigm. These candidate models are rule-based systems which mimic human-like reasoning according to principles of rational action (HumanToM) and ablations which do not factor in particular mental-states. We omit candidate models which make the same predictions as those listed in Table 1B.

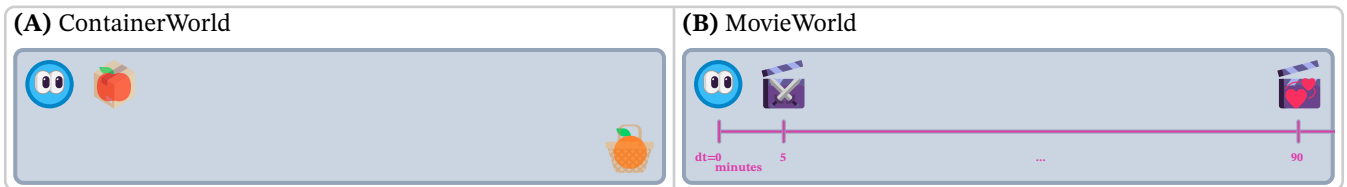

Figure 2: We transcribed each paradigm into a user prompt, requiring that GPT-4o produce a response with behavioral predictions or mental-state inferences, depending on the task. (A) An instance of the *ContainerWorld* paradigm with an apple in the box and an orange in the basket. The containers depicted are translucent for the purposes of this figure, but participants were told that the character was not able to see into them directly and instead had to make decisions based on prior beliefs about the contents of the boxes. (B) Similarly, in an instance of *MovieWorld*, the character would know that there is a movie starting in 5 minutes and a movie starting in 90 minutes, but would be unaware of the genre of each movie.

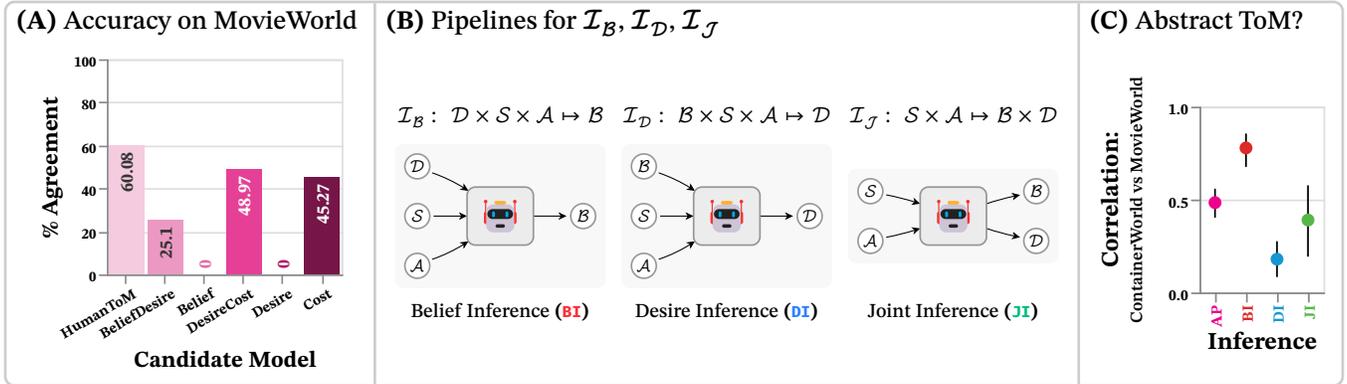

Figure 3: (A) GPT-4o's agreement compared with candidate models listed in Table 1B when tested on *MovieWorld* in Study 2. (B) When testing the LLM's mental-state inferences, we provided the model with a subset of the information contained in $\mathcal{B}, \mathcal{D}, \mathcal{S}, \mathcal{A}$, and queried it for belief inferences $\mathcal{I}_\mathcal{B}$, desire inferences $\mathcal{I}_\mathcal{D}$, or joint inferences $\mathcal{I}_\mathcal{J}$ such that the LLM had to simultaneously infer the character's beliefs and desires. (C) The correlation of GPT-4o's action predictions (AP; Study 2) and mental-state inferences across logically equivalent domains (*ContainerWorld* and *MovieWorld* Study 3).

While this approach allows us to capture an LLM's agreement with coherent ToM judgments, it is possible that LLMs might exhibit systematic limits relative to one dimension of their causal model. For example, judgments related to cost (implemented in *ContainerWorld* as physical distance) might prove challenging for disembodied systems known to have difficulties with physical reasoning (Webb et al., 2022). Thus, we also tested ablated candidate models, which remove each of the components of the full human ToM model (Table 1B).

**LLM Used** We evaluated GPT-4o via the OpenAI API. We left parameters at their default values: `temperature=1` and `top_p=1`. To push GPT-4o towards generating JSON responses, we set `response_format=json`. Evaluations reported in this work are from November 2024 using `gpt-4o-2024-05-13`.

**Results and Discussion**

Using the action predictions of the LLM, we can evaluate their agreement with the predictions from each candidate model. If an LLM has a "human-like ToM," then we would anticipate highest agreement with our HumanToM model, which incorporates the agent's desires, costs, and beliefs when making its action predictions. Indeed, we find that GPT-4o's responses have the highest coherence with HumanToM, providing congruent evidence with past claims that LLMs appear to have a human-like ToM.

While we found that HumanToM best aligns with GPT-4o's action predictions, we note that *ContainerWorld* is quite straightforward. For example, even under a human-like ToM, 85% of action predictions are to open the "box". This is because incurring the cost to open the "basket" (i.e., crossing the room) is only preferable when the character believes that the "box" exclusively contains their disliked fruit. However, even the much simpler Cost model (which always predicts that the character should open the "box") underperformed relative to HumanToM, suggesting that the simplicity of our paradigm did not prevent us from sufficiently differentiating between the candidate models tested here. Indeed, the margin between GPT-4o's agreement with Cost points to a more complex action-prediction mechanism as there are cases where GPT-4o correctly predicts the character should cross the room and take from the contents of the "basket". By this account, it appears that GPT-4o follows some core principles of action prediction – thus it may have a coherent ToM.

## Study 2: Is LLM ToM abstract?

A coherent relationship between costs, beliefs, and desires should not just apply to physical movements and the contents of containers. Instead, a ToM should apply across many different kinds of social situations even when superficial features vary greatly. We refer to this flexible use of a causal model of mental-states and behavior as "abstractness". For LLMs (and artificial agents more broadly), this abstractness has significant practical utility: if we want artificial agents that can navigate the social world, they must be equally good at understanding and predicting behaviors across myriad domains.

To this end, we developed an alternative paradigm, which we can use in conjunction with *ContainerWorld* to evaluate GPT-4o on its ability to replicate agreement (qualitatively and quantitatively) on behavioral predictions. Then, with the underlying distribution over actions (provided by GPT-4o), we contrast its behavioral predictions and mental-state inferences between both domains.

**Stimuli**

To evaluate a presence of a ToM across varied social scenarios, we needed a paradigm that (1) alters the underlying cost structure to ensure that LLMs success is not due to superfi-cial features, (2) every $\langle \mathcal{B}, \mathcal{D}, \mathcal{S} \rangle$ in *ContainerWorld* can be mapped onto this world (a 1:1 mapping), and (3) this paradigm should support a similarly comprehensive evaluation of both behavioral predictions ($\mathcal{F}$) and mental-state inferences ($\mathcal{I}_\mathcal{B}, \mathcal{I}_\mathcal{D}, \mathcal{I}_\mathcal{J}$), like *ContainerWorld* affords.

**Paradigm: *MovieWorld*** In our equivalent paradigm (Figure 2B), our character is at a foreign film festival – they want to see movies, but they have difficulty communicating with others because they do not speak the local language. At this film festival, there are two screenings coming up: one in 5 minutes and another in 90 minutes. Movies screened at this festival are 120 minutes long. The movies' genres are $\mathcal{S} \in$ {action, romance, action-romance}. The character also has beliefs about each screening's genre $\mathcal{B} \in$ {action, romance, action-romance}. The character's desires towards each genre may be either $\mathcal{D} \in$ {like, dislike}, but they may not "dislike" both. Lastly, the character may take an action $\mathcal{A} \in$ {5min, 90min} – this entails going to a screening and watching it.

## Does GPT-4o make the same behavioral predictions across domains?

Since every behavior in *ContainerWorld* has a corresponding behavior in *MovieWorld*, we anticipated that GPT-4o would produce the same pattern of responses across our paradigms.

**Procedure** To evaluate the similarity of predictions across paradigms, we replicated Study 1 in *MovieWorld*. First, we queried GPT-4o on all possible parameters combinations for *MovieWorld* to extract its forward model, $\mathcal{F}$. We then evaluated GPT-4o's agreement in *MovieWorld* against the candidate models previously described (Table 1B). These candidate models exhibit the exact same pattern of predictions across *ContainerWorld* and *MovieWorld* (e.g., where the closer location, "box," would have been predicted in *ContainerWorld*, "5 min" will be predicted in *MovieWorld*).

**Results** As in Study 1, we evaluated GPT-4o's agreement with our 6 candidate models (Figure 3A). Similar to its agreement results in *ContainerWorld*, GPT-4o obtained the highest agreement with `HumanToM`. However, the second highest agreement is with `DesireCost`, deviating from the patterns in *ContainerWorld* – this favors the idea that LLMs are attending to something more complex than "cost", but still markedly less complex than beliefs. Moreover, GPT-4o's agreement with these candidate models was notably attenuated in *MovieWorld* relative to *ContainerWorld* (Figure 3A). This likely stems from the change in cost representation – in *ContainerWorld* cost is an effort-based measure while in *MovieWorld* cost is a time-based measure. Although these results support show that LLMs' action predictions generally cohere with human-like models of ToM, it does so while converging with critical insights from Ullman (2023) – the ToM-like behavior of LLMs is fragile.

Our analysis thus far presupposes that LLMs, like GPT-4o, have a human-like (or proto-human-like) ToM. However, we aim to evaluate the flexibility of LLM ToM, divorced from its proto-human-like appearance. A more holistic measure would be evaluating if the behavioral predictions in *ContainerWorld* provide insight into the predictions in *MovieWorld*. We evaluate GPT-4o under this lens by using its action distribution in *ContainerWorld* to predict its action distribution in *MovieWorld* – as one would expect from a causal model. As Figure 3C (AP) depicts, though, *ContainerWorld* does not reliably predict *MovieWorld* ($r = .48, \text{CI}_{95\%}[0.41, 0.56]$).

## Are mental-state inferences abstract?

As ToM is a causal model of how mental-states generate behavior, this licenses us to infer an other's mental states from the behaviors we observe them engaging in. In our paradigms, this means that we can provide an observed behavior (an action) and some portion of mental-states (e.g., desires) and query an LLM to infer the missing mental-states (e.g., beliefs). We do this in both paradigms by querying an LLM to produce distributional estimates for each inferred mental-state (depicted in Figure 3B).

**Procedure** More concretely, for belief inferences, $\mathcal{I}_\mathcal{B}$, this entails generating each tuple of $\langle \mathcal{D}, \mathcal{S}, \mathcal{A} \rangle$ (there are $3 \times 9 \times 2$ such tuples for $\mathcal{I}_\mathcal{B}$) to synthesize parametrically varying `user` prompts. As when enumerating $\mathcal{F}$, we end the `user` prompt with a question for GPT-4o to generate its response ("What are Jason's beliefs?"). Similarly, we include a `system` prompt to inform GPT-4o of its task and the relevant JSON schema (all `system` and `user` prompts are available on OSF). We then extract the probability distribution over next tokens (e.g., "likes" apples and "likes" oranges), forming a distribution over desires $\mathcal{D}$ for each fruit (in *ContainerWorld*) or each genre (in *MovieWorld*).

We repeat similar processes for desire inferences, $\mathcal{I}_\mathcal{D}$, (with $9 \times 9 \times 2$ $\langle \mathcal{B}, \mathcal{S}, \mathcal{A} \rangle$ tuples) and for the joint belief-desire inference, $\mathcal{I}_\mathcal{J}$, (with $9 \times 2$ $\langle \mathcal{S}, \mathcal{A} \rangle$ tuples). With these estimates in-hand, we can use them to discern if each $\mathcal{I}_\mathcal{B}, \mathcal{I}_\mathcal{D}, \mathcal{I}_\mathcal{J}$ distribution from *ContainerWorld* predicts the logically corresponding distribution in *MovieWorld*.

**Results** When using the mental-state inferences from *ContainerWorld* to predict the inferences in *MovieWorld* (Figure 3C), we find that only beliefs are able to reliably predict each other across our two paradigms $r = .78, \text{CI}_{95\%}[.68, .85]$. While $\mathcal{I}_\mathcal{B}$ are strongly associated across paradigms, $\mathcal{I}_\mathcal{J}$ is only weakly associated across paradigms ($r = .39, \text{CI}_{95\%}[.2, .57]$) and $\mathcal{I}_\mathcal{D}$ is very weakly associated across paradigms ($r = .18, \text{CI}_{95\%}[.09, .27]$). Perhaps the $\mathcal{I}_\mathcal{B}$ resembles some form of an abstract representation, but in-aggregate GPT-4o does not possess an abstract representation of mental-states.

## Discussion

An abstract ToM should be qualitatively and quantitatively similar across 1:1 domains. However, we find that LLMs make substantively different behavioral predictions ($\mathcal{F}$) and mental-state inferences ($\mathcal{I}$) across our linked paradigms.

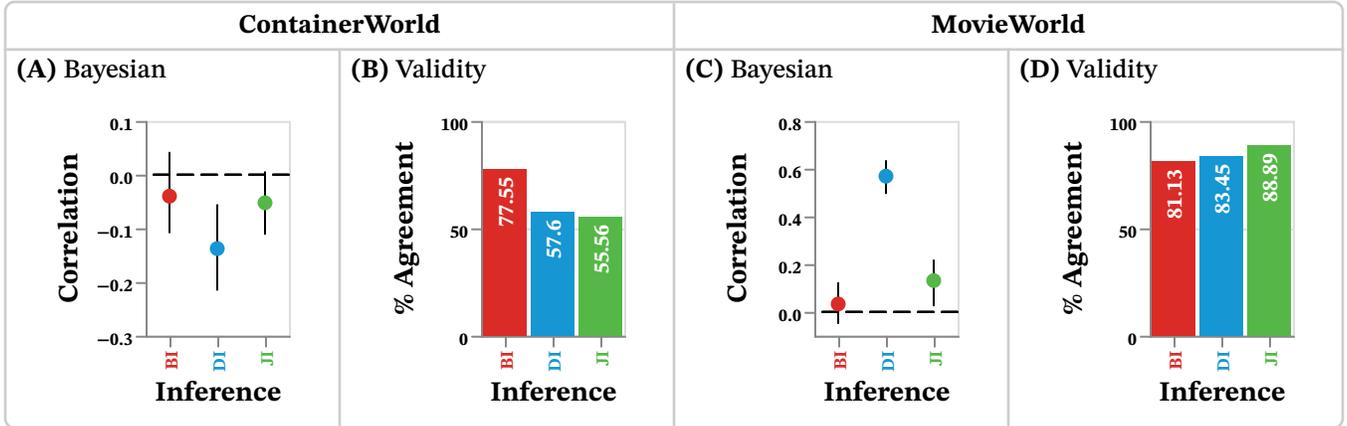

Figure 4: (A) GPT-4o's correlation for each mental-state inference (beliefs $\mathcal{I}_\mathcal{B}$, desires $\mathcal{I}_\mathcal{D}$, and joint belief-desire $\mathcal{I}_\mathcal{J}$) between the expected posterior (from $\mathcal{F}$) and GPT-4o's direction estimates for each mental-state ($\mathcal{I}_\mathcal{B}$, $\mathcal{I}_\mathcal{D}$, or $\mathcal{I}_\mathcal{J}$) in *ContainerWorld*. (B) GPT-4o's accuracy of inferring mental-states ($\mathcal{I}_\mathcal{B}$, $\mathcal{I}_\mathcal{D}$, and $\mathcal{I}_\mathcal{J}$) that can be used as input to $\mathcal{F}$ and generate the observed behavior $\mathcal{A}$ in *ContainerWorld*. (C and D) Repeats these measurements, but in *MovieWorld*.

Qualitatively, our results in the $\mathcal{F}$ (Figures 1C and 3A) spark an important caveat to Study 1: on any single task, LLMs may demonstrate high agreement with human judgments – as we saw in both paradigms. However, these judgments are brittle and may not be equally applicable across similar domains. As ToM is a domain-general causal model of how mental-states generate behavior, we would expect that both behavioral predictions and mental-state inferences would predict each other across our paradigms (Figure 3C). In the most ideal scenario, these correlations would be at, or near, ceiling. However, we ultimately found that only belief-inferences were able to reliably predict direct estimates across paradigms.

Taken together, this suggests that while LLMs, like GPT-4o are capable of generate reasonable behavioral predictions and inferences, these inferences, in-aggregate, do not seem to be generated by an abstract model of other minds – a core requirement of ToM.

## Study 3: Is LLM ToM consistent?

One possible explanation for why LLMs might have a coherent, but not abstract ToM is that LLMs might synthesize narrow causal models within certain domains. Thus, they may have many ToMs, which could individually cohere with human judgments (as seen in Study 1), but do not cohere with each other across different domains (Study 2). To test this explanation, we turn our attention to the final core feature of ToM: consistency. This core feature specifies that the causal model underpinning a ToM is causally linked. That is, in either direction of ToM, we expect the $\langle \mathcal{B}, \mathcal{D}, \mathcal{S}, \mathcal{A} \rangle$ to be the same, even if predictions might be "wrong". For instance, consider Figure 2B: if the character attends the "90min" screening and believes it is a romance movie, but a ToM infers the character "likes" action movies, as long as the same desire towards action movies generates the "90min" action while believing a romance movie will be shown, then this ToM is consistent.

Thus, we can evaluate whether LLMs have anything like a ToM – even ones that fail to meet the criteria of abstractness – by testing whether their behavioral predictions are derived from their mental-state inferences (and likewise, in reverse).

### Procedure

To evaluate the consistency of LLM ToM, we tested for prediction-inference agreement across the three mental-state inference tasks: $\mathcal{I}_\mathcal{B}$, $\mathcal{I}_\mathcal{D}$, and $\mathcal{I}_\mathcal{J}$ in both paradigms (*ContainerWorld* and *MovieWorld*). First, we computed the predicted inferences under a Bayesian inversion of $\mathcal{F}$ to get the expected posterior (Baker et al., 2017). Then we correlated this posterior to the likelihood generated by the LLM (Figure 3B) – hereafter referred to as the "Bayesian evaluation".

This Bayesian evaluation, though, demands that LLMs leverage a Bayesian ToM, meaning that the probability of a mental-state $m$ is proportional to probability of the action $a$ and other mental-states $n$ – more formally: $P(m \mid a, n) \propto P(a, n \mid m) \cdot P(m)$. Since this may not be the case, we also consider a more generous evaluation metric based on the internal validity of the inputs and outputs of the LLM when performing action prediction and inference over corresponding tuples, which we refer to as our "validity evaluation". Crucially, this metric requires that a mental-state inference, when used as input to $\mathcal{F}$, produces the target action to be explained, a requirement of any causal model. This makes for a more generous metric because there are many $\mathcal{B}, \mathcal{D}$ inputs to $\mathcal{F}$ which will generate the target action $\mathcal{A}$.

For both of these evaluations, we would anticipate correlations to be at, or near, ceiling.

### Results and Discussion

Because our "Bayesian evaluation" is a stricter metric, either at or near ceiling correlations across both paradigms would illustrate a consistent ToM. Instead, we find that across both paradigms, GPT-4o fails on both of these criteria (Figures 4A and 4C). Our "validity evaluation" is a more forgiving evaluation: many mental-states could be inferred from an observed action, but we only require that one of these inferred mental-states generates the target action. Because of this, we hold GPT-4o to a higher threshold of ceiling agreement. Under this metric we find that across both paradigms GPT-4o fails to reach ceiling (Figures 4B and 4D). Taken together, these failures illustrate that GPT-4o's action predictions (from mental-states) are unrelated to its mental-state inferences (from actions).

When evaluating GPT-4o for consistency in its forward and backward models, we find that across two metrics – a Bayesian inversion and a more generous "are inferred mental-states able to generate observed behavior?" – GPT-4o fails to retain a consistent ToM. As such, GPT-4o lacks the final core component of a ToM: consistency.

## General Discussion

Humans have to navigate a variety of complex social situations on a daily basis. Regardless of whether we are reasoning about what snack our picky toddler might choose at a birthday party or the beliefs of a friend who wore a vampire costume to a fancy dinner, our ToM enables us to have some predictive power over the chaotic social world. Our coherent, abstract, and consistent causal model of mental states and behaviors constitutes an invaluable generative social cipher.

LLMs, however, seem to achieve their social proficiency in the absence of a causal model. While LLMs are capable of making predictions about others' actions in a way that largely coheres with the principles of human action prediction (Study 1; e.g., Kosinski (2024)), this coherence is brittle as LLM's predictive success decreases significantly when applied to a different but logically equivalent domain (Study 2). Finally, we found that even within a single domain, LLMs fail to generate reciprocal action predictions and inferences for a given scenario (Study 3). This contrasts with the representations posited in ToM, which involve a single causal model that is used to both predict and interpret the behavior of others. Taken together, our investigation suggests that current LLMs do not possess unified abstract principles regarding the relationship between mental states and behavior. Thus, this work should cast serious doubt on the prospect of LLM ToM.

While we failed to find evidence of LLM ToM, our cognitively-grounded approach is actually more charitable than previous benchmark evaluations of ToM (e.g., Trott et al. (2023)). Specifically, our abstractness and consistency evaluations (Studies 2 and 3) remove the implicit expectation that LLM ToM must be human-like. Indeed, there is not one ToM. For example, ToM causal models differ between children and adults (Onishi & Baillargeon, 2005; Wellman & Liu, 2004), between human and non-human primates (Martin & Santos, 2016; Rosati et al., 2010), and exhibit some variability across cultures (Liu et al., 2008; Yu & Wellman, 2024). In the same way, LLMs might have their own emergent ToM – one that differs from human ToM and therefore might be missed by benchmark evaluations which typically use adult human responses as ground-truth (e.g., Strachan et al. (2024)). Thus, our investigation intentionally avoided human-model evaluations as the benchmark for a "presence of ToM". However, even under our less anthropocentric evaluations, our target LLM still failed to reliably perform across domains (Study 2) and did not generate internally consistent action predictions and mental-state inferences (Study 3).

One limitation of the current investigation is that we only evaluate one LLM with our proposal. We specifically chose GPT-4o as our target system since it was regarded as the most advanced and widely available LLM at the time. While we find that GPT-4o lacks a ToM, there remains an open question: is this restricted to GPT-4o or generalizable across LLMs? We would expect other LLMs to perform similarly because they struggle to recover causal models from statistical regularity (Vafa et al., 2024). Of course, it is possible that future, higher-parameter LLMs trained on larger datasets could develop an emergent ToM. However, it is also possible that these future LLMs will instead continue to increase in social proficiency without ever having a ToM. Thus, on the precipice of increasingly powerful LLMs, evaluations like ours that prioritize the relevant features of a capacity, rather than human-like benchmarking will be of increasing importance. To aid computational researchers, we also plan to formalize this approach as an open-source evaluation metric that will allow them to assess the coherence, abstractness, and consistency of their model's representation of other minds.

The failures exhibited by GPT-4o, however, raise another important question: Should we care whether or not LLMs have a ToM? We would like to argue that the answer is yes; it does matter whether LLMs pass benchmarks (e.g., Kosinski (2024)) with or without the benefit of a ToM. Claiming that an LLM possesses a ToM implies that it can make reasonable, coherent mental state judgments even when asked to generalize far outside of its training data. Thus, if an LLM had a ToM, we would be able to have more confidence in the "reasonableness" of its outputs even across widely varying social situations.

Beyond its application to future LLMs, we hope that the spirit of this evaluation can also be applied to other intelligent systems and capacities. For example, nonhuman primates, much like LLMs, demonstrate notable social proficiency. However, debates regarding their ToM focus on the kinds of mental state representations that nonhuman primates can hold (Martin & Santos, 2016; Rosati et al., 2010) without discussing whether their ToM-like capacities are coherent, abstract, and consistent. Additionally, ToM is just one of many folk theories, like physics, sociology, and economics – all of which posit a causal generative model. In future work, we plan to explore the existence of other intuitive theories within LLMs. We hope that a cognitively-grounded approach can both enable clearer, more meaningful desiderata for success and even allow us to recognize diverse intelligences with causal models that look very different from our own.

## Acknowledgements

This work was supported by NSF award IIS-2106690.

## Code and Data Availability

Code and data for this paper can be found at the accompanying OSF repository: `https://osf.io/8eha3/`.